\documentclass[conference]{IEEEtran}
\IEEEoverridecommandlockouts

\usepackage{cite}
\usepackage{amsmath,amssymb,amsfonts}
\usepackage{graphicx}
\usepackage{textcomp}
\usepackage{booktabs}
\usepackage{multirow}
\usepackage{url}
\usepackage[hidelinks]{hyperref}

\AtBeginDocument{\makeatletter\def\@cite#1#2{\textsuperscript{[{#1\if@tempswa , #2\fi}]}}\makeatother}

\begin{document}

\title{Support Thresholds, Not Algorithms, Limit\\Rare-Association Recovery in Co-Purchase Networks}

\author{\IEEEauthorblockN{1\textsuperscript{st} Xiao Han*}
\IEEEauthorblockA{\textit{Goizueta Business School} \\
\textit{Emory University}\\
Atlanta, GA, USA \\
xhan@alumni.emory.edu}
\and
\IEEEauthorblockN{2\textsuperscript{nd} Zhen Zhang}
\IEEEauthorblockA{\textit{School of Data Science} \\
\textit{University of Pennsylvania}\\
Philadelphia, PA, USA \\
billzhangzhen98@gmail.com}
\and
\IEEEauthorblockN{3\textsuperscript{rd} Xin Zhao}
\IEEEauthorblockA{\textit{Olin Business School} \\
\textit{Washington Univ.\ in St.\ Louis}\\
Saint Louis, MO, USA \\
maxizhao9@gmail.com}
\and[\\\hfill]
\IEEEauthorblockN{4\textsuperscript{th} Jiechun Lei}
\IEEEauthorblockA{\textit{Independent Researcher} \\
\textit{Independent Researcher}\\
New Haven, CT, USA \\
chloe.lei@aya.yale.edu}
\and
\IEEEauthorblockN{5\textsuperscript{th} Moxuan Zheng}
\IEEEauthorblockA{\textit{Stern School of Business} \\
\textit{New York University}\\
New York, NY, USA \\
mz2156@nyu.edu}
\and
\IEEEauthorblockN{6\textsuperscript{th} Youting Wang}
\IEEEauthorblockA{\textit{College of Engineering} \\
\textit{Northeastern University}\\
Mountain View, CA, USA \\
ginkoin613@gmail.com}
}

\maketitle

\begin{abstract}
The support threshold of the Apriori algorithm involves a trade-off in conducting market basket analysis: the associations that occur frequently are noted with high threshold; however, the low ones lead to generating the large amount of rules.
The paper compares five methods for co-purchase edge filtration on two grocery datasets: i.e., Instacart (3.2 million baskets) and Dunnhumby (208 thousand baskets), including Apriori, Apriori + lift post-filtering, top-$K$ ranking based on lift, and two methods based on networks, noise-corrected (NC) and disparity filter (DF).
The top-$K$ method ensures the maximum average lift, while the NC achieves similar lift level by means of a single value of the significance parameter~($\alpha$).
These two methods recover substantially more rare high-lift associations than Apriori (80--100\% against 22--28\%).
NC and top-$K$ select meaningfully different edges (18--29\% non-overlapping): NC retains statistically validated pairs, while top-$K$ retains rare pairs with high lift but low statistical significance.
A rolling-origin holdout evaluation shows that top-$K$ edges recur at higher rates at every split, but NC edges are ${\sim}12$~pp more likely to remain statistically significant in the held-out network.
\end{abstract}


\section{Introduction}

Introduced by Agrawal and Srikant in 1994, Association rule mining (ARM) has laid an important foundation in market basket analysis, particularly through the Apriori algorithm~\cite{agrawal1994}.
Apriori has minimum standards of support and confidence for generating rules, and high levels of support ensure that rules produced are meaningful but at the expense of infrequent rules; similarly, when support standards are low one gets too many rules owing to combinatorial explosion of rules~\cite{tan2004, geng2006}.
The problem thus still persists as algorithms used to generate rules, including, but not limited to FP-Growth~\cite{han2000} and Eclat~\cite{zaki2000} still rely on minimum level of support being met before rules are visible.
In this paper, we restate the problem of rare associations as one of \emph{visibility}.
The problem is not that of the sophistication of the algorithm, but the minimum level of support which has to be established before the rule becomes visible.
Any approach that circumvents this barrier, be it based on statistics or non-statistics, is able to recover rare association rules, which have thus far remained undiscovered by application of Apriori algorithm.
Similar challenges of visibility also exist in the domain of fraud detection, where due to extremely unequal class distribution rare patterns or instances become invisible to threshold-based algorithms~\cite{sun2025fraud, sun2025gru}.

Network-based weight data on retail co-purchases is available.
Prior studies used co-purchase graphs for community detection~\cite{hsieh2022, gao2023}, network visualization~\cite{gino2023}, and product association mining~\cite{kholod2024, wahidi2024}, but without edge-significance testing.
Backbone extraction methods, the disparity filter (DF)~\cite{serrano2009} and noise-corrected model (NC)~\cite{coscia2017}, retain edges only when the observed weight is statistically significant under a null model.
These have been applied to trade, airline and collaboration networks~\cite{serrano2009, coscia2017, yassin2025comparison}, but never to co-purchase networks.
The closest related work includes Tian et al.~\cite{tian2021} on complement/substitute relations and Musciotto et al.~\cite{musciotto2022} on higher-order simplices in Walmart data.

We present three contributions as follows:
(1)~without a global support floor, both top-$K$ and NC recover 80--100\% of rare high-lift edges (NC achieves this through a genuine significance test); Apriori+lift post-filtering cannot compensate because the support floor limits the candidate pool before lift ranking takes effect;
(2)~even with a more or less similar lift measure of the edges recovered, NC and top-$K$ produce completely different edge sets (18--29\% do not overlap) because NC evaluates statistically significant pairs and such significance persists on held data (${\sim}12$~pp higher persistence of significance), while around 89\% to 93\% of pairs returned by the top-$K$ method show negative $z$-scores in the training and the test networks, which correlate with the well-known fact of lift being inflated at the very low support~\cite{tan2004};
and (3)~we found out that the DF cannot be applied universally, as in particular its within-node null does not align with between-node lift in the co-purchasing networks.

\section{Method}

\subsection{Co-Purchase Network Construction}

Given shopping baskets, we construct a weighted undirected graph $G = (V, E, w)$ where nodes are product categories and edge weight $w_{ij}$ equals the number of baskets containing both $i$ and $j$, retaining only edges with $w_{ij} \geq 2$ to suppress singleton co-occurrences.
For each edge we compute lift: $\text{lift}(i,j) = \text{support}(i,j) / [\text{support}(i) \cdot \text{support}(j)]$.

\subsection{Backbone Extraction Methods}

\textbf{Disparity Filter (DF)}~\cite{serrano2009}.
For node $i$ with degree $k_i$ and strength $s_i$, the DF tests whether normalized weight $p_{ij} = w_{ij}/s_i$ exceeds a uniform null.
The $p$-value is $\alpha_{ij} = 1 - (k_i - 1) \int_0^{p_{ij}} (1-x)^{k_i - 2} dx$; an edge is retained if $\alpha_{ij} < \alpha$ from either endpoint.

\textbf{Noise-Corrected Model (NC)}~\cite{coscia2017}.
We use the binomial-null formulation (default in reference toolkits~\cite{yassin2023netbone, neal2022}).
Expected weight: $\mu_{ij} = s_i s_j / 2W$ where $W = \frac{1}{2}\sum_i s_i$.
Variance: $\sigma^2_{ij} = \mu_{ij}(s_i + s_j - 2\mu_{ij})/(2W-1)$.
A $z$-score $z_{ij} = (w_{ij} - \mu_{ij})/\sigma_{ij}$ yields a one-sided $p$-value; edges with $p < \alpha$ are retained.
The NC null considers \emph{both} endpoints' strength simultaneously: $\mu_{ij} \propto s_i \cdot s_j$, mirroring lift's denominator $\text{support}(i) \cdot \text{support}(j)$.
This structural alignment means edges that are surprising under NC's null tend to have high lift, explaining their empirical correlation.

\subsection{Baselines}

\textbf{Apriori}~\cite{agrawal1994}: run via \texttt{mlxtend} at 7 support thresholds ($s \in \{0.001, 0.005, 0.01, 0.02, 0.05, 0.1, 0.2\}$) $\times$ 5 confidence thresholds ($c \in \{0.01, 0.05, 0.1, 0.2, 0.5\}$), restricted to pairs.
\textbf{Top-$K$-by-Lift}: all edges sorted by lift, top $K$ retained; by construction this maximizes average lift for any $K$-sized subset, making it an upper-bound diagnostic, not a production method.
\textbf{Apriori+Lift Filter}: Apriori at low support, top $K$ by lift retained.
All backbone methods implemented in Python from original formulations~\cite{serrano2009, coscia2017}.

\subsection{Rare-Rule Recovery}
We define \emph{rare high-lift associations} operationally as edges with $s < 0.01$ and lift above 2.
At Apriori's lowest tested threshold ($s = 0.001$), associations in the $[0.001, 0.01)$ support band enter the candidate set, but the majority of rare associations---those below $s = 0.001$, remain structurally excluded.
For each method, rare-rule recovery is the fraction of all such edges present in the method's retained set.

\section{Experiments}

\subsection{Datasets}

\begin{table}[t]
\centering
\caption{Dataset and network statistics.}
\label{tab:datasets}
\begin{tabular}{@{}lrr@{}}
\toprule
& \textbf{Instacart} & \textbf{Dunnhumby} \\
\midrule
Baskets & 3,182,490 & 208,119 \\
Categories & 134 (aisle) & 303 (commodity) \\
Network edges & 8,910 & 37,667 \\
Network density & 0.999 & 0.823 \\
Mean lift (full) & 1.50 & 2.04 \\
\bottomrule
\end{tabular}
\end{table}

We evaluate on two publicly available grocery datasets (Table~\ref{tab:datasets}).
\textbf{Instacart}~\cite{instacart2017}: 3.2M baskets across 134 aisles yield a near-complete co-purchase graph (density 0.999).
\textbf{Dunnhumby}~\cite{dunnhumby2017}: 208K baskets across 303 commodity categories yield a sparser graph (density 0.823).

\subsection{Evaluation Metrics}
We report mean and median lift, fraction of edges with lift above 2, cross-department diversity, rare-rule recovery (Section~II-D), and holdout recurrence (Section~\ref{sec:holdout}).
$K$ equals NC's edge count at $\alpha = 0.01$, fixed following standard backbone extraction practice~\cite{yassin2025comparison, neal2022} with no dataset-specific tuning; Apriori is tested at its best of 35 support$\times$confidence configurations.

\subsection{Main Results}

\begin{table*}[t]
\centering
\caption{Best-configuration comparison. $K$ is set to NC's edge count at $\alpha{=}0.01$; no parameter was tuned to advantage any method. Bold = best per dataset; $\dagger$ = upper-bound diagnostic (not a deployable method); $*$ = Apriori candidate pool exhausted ($<K$ edges available at $s{=}0.001$).}
\label{tab:main}
\begin{tabular}{@{}llrrrrrr@{}}
\toprule
\textbf{Dataset} & \textbf{Method} & \textbf{Edges} & \textbf{Mean Lift} & \textbf{Med.\ Lift} & \textbf{Lift${>}$2 (\%)} & \textbf{Cross-Dept (\%)} & \textbf{Rare Rec.\ (\%)} \\
\midrule
\multirow{5}{*}{Instacart}
& Top-$K$$^\dagger$ ($K{=}3949$) & 3,949 & \textbf{2.05} & \textbf{1.63} & \textbf{24.1} & 87.8 & \textbf{100.0} \\
& NC ($\alpha{=}0.01$) & 3,949 & 2.01 & 1.62 & \textbf{24.1} & \textbf{87.8} & \textbf{100.0} \\
& Apr+Lift ($s{=}.001$) & 3,949 & 1.47 & 1.34 & 7.0 & 92.8 & 27.2 \\
& Apriori ($s{=}.001$) & 3,040 & 1.48 & 1.34 & 7.6 & 91.7 & 22.2 \\
& DF ($\alpha{=}0.20$) & 2,909 & 1.53 & 1.32 & 8.7 & 91.5 & 24.4 \\
\midrule
\multirow{5}{*}{Dunnhumby}
& Top-$K$$^\dagger$ ($K{=}16705$) & 16,705 & \textbf{3.07} & \textbf{2.32} & \textbf{75.1} & 72.1 & \textbf{100.0} \\
& NC ($\alpha{=}0.01$) & 16,705 & 2.88 & 2.24 & 60.4 & 73.3 & 80.5 \\
& Apr+Lift$^*$ ($s{=}.001$) & 11,490 & 2.03 & 1.89 & 42.9 & 74.2 & 34.9 \\
& Apriori ($s{=}.001$) & 8,324 & 2.16 & 2.00 & 49.7 & 70.8 & 28.2 \\
& DF ($\alpha{=}0.20$) & 13,554 & 2.15 & 1.81 & 39.7 & \textbf{77.9} & 38.8 \\
\bottomrule
\end{tabular}
\end{table*}

Table~\ref{tab:main} compares all five approaches.
Top-$K$ and NC score at least $1.4$--$1.5$ times higher mean lift than Apriori.
Importantly, even Apr+Lift, which uses lift after filtering the output of Apriori, loses to NC because the support floor restricts Apriori's candidates before lift can be applied.
The impact of this is quite large in Dunnhumby: at $s = 0.001$ with no confidence filter, Apriori's candidate pool contains only 11{,}490 edges (the Apr+Lift input in Table~\ref{tab:main}), while NC retains 16{,}705. Post hoc filtering cannot create what was never made.
The DF achieves a result similar to that of Apriori (1.53 / 2.15) due to a null-model mismatch: its within-node test does not match between-node lift in a dense co-purchase network~\cite{serrano2009}.
In terms of rare-rule recovery, NC achieves 80--100\% recoveries of rare high-lift edges in comparison with Apriori's 22--28\%, confirming that the support floor is the cause of the problem.

\begin{figure}[t]
\centering
\includegraphics[width=\columnwidth]{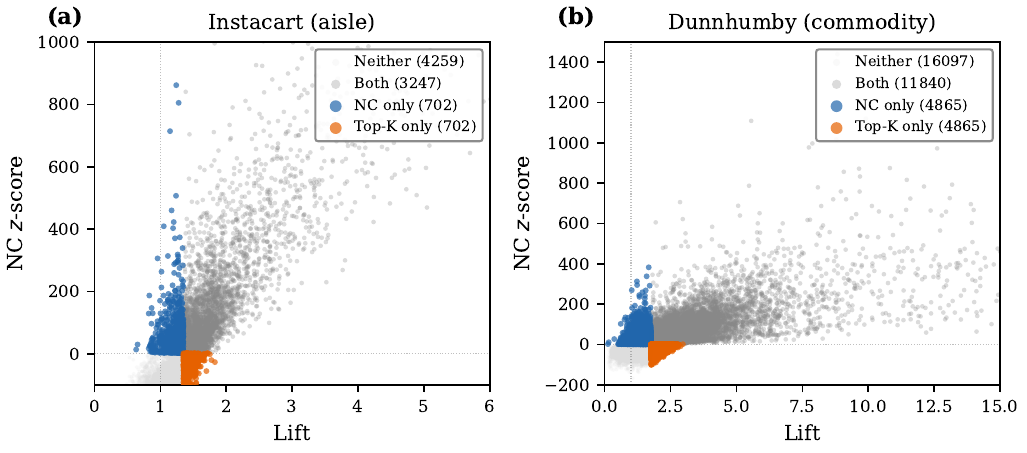}
\caption{Lift vs.\ $z$-score for each edge at $\alpha{=}0.01$ (matched to Table~\ref{tab:main}), colored by method selection. Upper-right: high lift and statistically significant (retained by both). Upper-left: high lift but negative $z$ (top-$K$-only; lift inflation). Lower-right: moderate lift but highly significant (NC-only; popular-category pairs).}
\label{fig:scatter}
\end{figure}

\textbf{Divergence at the edge set.}
At $\alpha = 0.01$, edges of NC and the top-$K$ overlap by 70--82\% (Jaccard values of 0.55--0.70), leaving 18--29\% of edges outside the NC-top-$K$ joint edge set.
DF overlaps less with both methods (Jaccard 0.25--0.31), consistent with its distinct within-node null.
Edges exclusive to NC have high $z$-scores (mean 65.9 vs.\ 25.3, far above the Bonferroni threshold $z \approx 5.0$ at $\alpha/|E|$) while lift values are acceptable (1.20 vs.\ 1.36), confirming statistically validated associations missed by lift alone.
Edges exclusive to top-$K$ will have high lift values (1.44 against 2.04) but low (negative) $z$-scores ($-36.8$ against $-19.0$).
The most negative-$z$ top-$K$-only edges are niche$\times$common pairs (e.g., ``dry dinner mixes $\times$ tomatoes,'' lift~$= 1.8$, $z = {-}95$) whose high lift reflects low marginal frequency rather than a genuinely surprising pattern.
Figure~\ref{fig:scatter} visualizes this separation: NC edges cluster at high $z$ with moderate lift, while top-$K$-only edges cluster at high lift with negative $z$, foreshadowing the temporal-stability findings in Section~\ref{sec:holdout}.

\begin{figure}[t]
\centering
\includegraphics[width=\columnwidth]{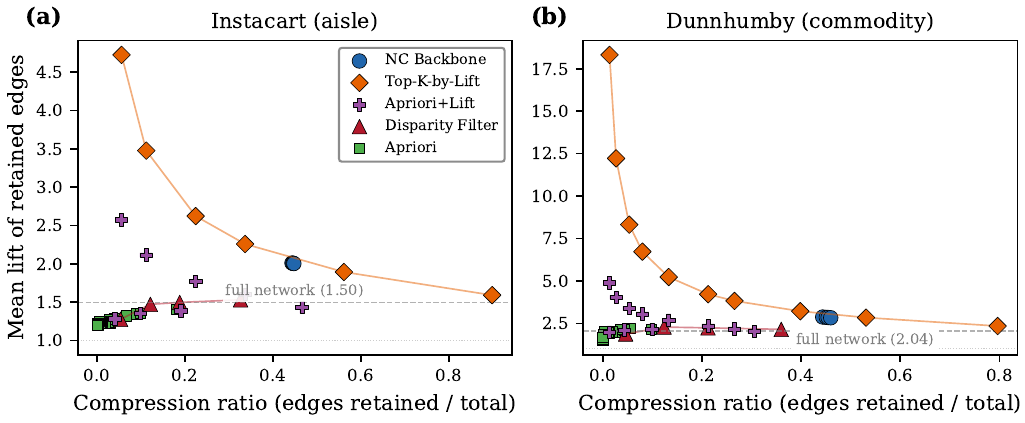}
\caption{Mean lift vs.\ compression ratio across multiple $\alpha$ and $K$ configurations. Top-$K$ and NC trace a Pareto frontier; DF clusters with Apriori variants. Table~\ref{tab:main} reports the $\alpha{=}0.01$ operating point.}
\label{fig:precision}
\end{figure}

\textbf{Precision at density} (Figure~\ref{fig:precision}).
Top-$K$ and NC trace a joint Pareto frontier above all other methods at every compression level; NC performance varies smoothly across $\alpha$ values with no abrupt threshold effect.

\subsection{Temporal Holdout Evaluation}
\label{sec:holdout}

Metrics that are based on lift are ineffective in differentiating between genuine permanent relationships and temporary associations.
We apply rolling-origin evaluation at five train splits (50--90\%) and construct separate co-purchase networks from training and test transactions to measure co-occurrence stability, not conversion or recommendation accuracy.

On Instacart, all methods achieve 100\% recurrence (near-complete graph), but NC edges average $1.5\times$ more test co-occurrences than top-$K$ and ${\sim}12$~pp higher significance persistence across all splits.

Dunnhumby (Table~\ref{tab:rolling}) demonstrates that top-$K$ presents better recurrence than NC at all five splits (0.5--6.7~pp gap), while NC presents better significance persistence at all five splits (8.3--14.2~pp gap).
The two metrics highlight different qualities; recurrence seeks to establish occurrence of the pair, whereas significance persistence tests whether the pair occurred more frequently than null hypothesis posits.
This distinction is significant as 89 to 93\% of edges discovered by top-$K$ only are associated with negative $z$-scores in both training and test datasets, indicating their co-occurrence is lower than expected according to the overall connectivity of the given categories meaning that the high lift of the pairs results from their low marginal frequency rather than true surprising co-occurrence.
NC eliminates these statistically inconclusive edges while preserving the associations with the temporally stabilized significance.
DF indicates nearly perfect recurrence ($99.6\%$); significance persistence is omitted because only 44--54\% of DF edges are NC-significant even in the full network, so the NC-based metric is uninformative for this method.

\begin{table}[t]
\centering
\caption{Rolling-origin holdout on Dunnhumby. Recurrence: fraction of retained edges co-occurring $\geq 1$ time in test. Sig.\ persist.: fraction remaining significant ($p{<}0.01$) in the test-period network. Top-$K$ leads on recurrence; NC leads on significance persistence.}
\label{tab:rolling}
\begin{tabular}{@{}ccc|cc|c@{}}
\toprule
& \multicolumn{2}{c}{\textbf{Recurrence (\%)}} & \multicolumn{2}{c}{\textbf{Sig.\ Persist.\ (\%)}} & \textbf{DF} \\
\textbf{Train} & \textbf{NC} & \textbf{Top-$K$} & \textbf{NC} & \textbf{Top-$K$} & \textbf{Rec.} \\
\midrule
50\% & 97.1 & \textbf{97.6} & \textbf{70.3} & 56.1 & 100.0 \\
60\% & 95.0 & \textbf{96.2} & \textbf{68.4} & 54.1 & 100.0 \\
70\% & 91.8 & \textbf{93.9} & \textbf{65.1} & 51.8 & 100.0 \\
80\% & 87.1 & \textbf{90.9} & \textbf{60.0} & 48.6 & 99.6 \\
90\% & 78.3 & \textbf{85.0} & \textbf{51.7} & 43.4 & 98.5 \\
\midrule
\textbf{Mean} & 89.9 & \textbf{92.7} & \textbf{63.1} & 50.8 & 99.6 \\
\bottomrule
\end{tabular}
\end{table}

\section{Discussion}

The rare-association problem in market-basket analysis is fundamentally one of visibility at a given support threshold.
Both NC and top-$K$ bypass this threshold: on Dunnhumby, NC at $\alpha = 0.01$ yields 16{,}705 edges versus Apriori's candidate pool of 11{,}490 at $s = 0.001$ (no confidence filter), confirming the support floor as a structural impediment.

\textbf{Temporal stability.}
Top-$K$ produces a larger number of binary recurrence across the 5 splits but NC shows a much larger persistence of significance (8--14 pp per split across both datasets).
As evident from Dunnhumby, 89--93\% of edges identified only as top-$K$ edges have negative $z$-scores in both training and testing networks, and therefore one can conclude that high lift is due to low marginal frequency rather than unexpected co-occurrence.
While NC discards such edges and includes only the relationships which meet the statistical significance criterion through held-out datasets, which is more suitable for retail recommendation than just recurrence.

The DF achieves only Apriori-comparable lift because its within-node null is misaligned with between-node lift in dense co-purchase graphs~\cite{serrano2009, li2025catch}.

\textbf{Deployment perspective.}
The decision that the practitioner is facing is between NC and Apriori algorithm techniques rather than NC and top-$K$ method.
Since top-$K$ computes pairwise lift, it is a diagnostic limit that is not a filter.
The advantage of NC is its only understandable parameter~($\alpha$) and the fact that it generates connected subgraphs suitable for evaluating associations between items.
Therefore, if the practitioner needs a complete list of associations between items, top-$K$ should be invoked.
If the intention is to find surprising relationships among items, NC remains a better alternative.

\section{Conclusion}

This study showed that the support threshold, not the mining algorithm, limits rare-association recovery: NC backbone extraction with a single parameter recovers 80--100\% of rare high-lift edges whose significance persists in held-out data (8--14~pp above top-$K$).
Top-$K$'s superiority on lift-based metrics is partly definitional.
Both datasets are US grocery at category level; generalization to SKU-level networks, other verticals, or non-US markets is untested, and holdout recurrence measures co-occurrence stability, not conversion.
Future work includes SKU-level networks, non-grocery domains, coupling NC-filtered edges with recommendation models, and leveraging large language models for automated interpretation of discovered associations~\cite{liu2026llm}.

\end{document}